\documentclass[10pt,twocolumn,letterpaper]{article}

\usepackage[pagenumbers]{wacvTemplate/wacv} % To force page numbers, e.g. for an arXiv version

\usepackage{graphicx}
\usepackage{amsmath}
\usepackage{amssymb}
\usepackage{booktabs}
\usepackage[]{adjustbox}
\usepackage[percent]{overpic}
\usepackage{multicol}
\usepackage{stfloats}

\usepackage[dvipsnames]{xcolor}
\usepackage{colortbl}
\usepackage[]{todonotes}
\graphicspath{{figures/}}

\newcommand{\bftab}{\fontseries{b}\selectfont}
\newcommand{\deltaavg}{$<$$\delta^x_{avg}$}
\newcommand{\method}[1]{\mbox{\textsc{#1}}}
\newcommand{\dataset}[1]{\mbox{\textsc{#1}}}
\newcommand{\mat}[1]{{\mathbf{#1}}}
\newcommand{\vect}[1]{{\mathbf{#1}}}
\newcommand{\resdeltap}[1]{{\tiny ({\color{OliveGreen}$\mathbf{#1}$})}}
\newcommand{\resdeltan}[1]{{\tiny ({\color{BrickRed}$\mathbf{#1}$})}}
\newcommand{\frst}[1]{\bftab{#1}}
\newcommand{\scnd}[1]{\underline{#1}}

\newcommand{\ft}[2]{{#1 \rightarrow #2}}   % \fromto{0}{t} outputs 0 -> t

\newcommand{\sampleat}[2]{#1\left[#2\right]}
\newcommand{\stimes}{{\mkern-1mu\times\mkern-1mu}}
\DeclareMathSymbol{\shortminus}{\mathbin}{AMSa}{"39}

\newcommand{\Flow}{\ensuremath{\mathbf{F}}}
\newcommand{\Occl}{\ensuremath{\mathbf{O}}}

\newcommand{\Unc}{\ensuremath{\mathbf{U}}}
\newcommand{\Qualitymethod}{\mathcal{Q}}

\newcommand{\Img}{\ensuremath{I}}

\newcommand{\Deltaset}{\ensuremath{\mathcal{D}}}
\newcommand{\template}{template\xspace} % OR reference? \xspace for correct spaces after word

\newcommand{\Cost}{\ensuremath{\mathbf{E}}}
\newcommand{\Costname}{Cost\xspace}
\newcommand{\costname}{cost\xspace}

\usepackage{physics} % for \abs{} and \norm{}

\usepackage[pagebackref,breaklinks,colorlinks]{hyperref}
\usepackage{pdfcomment}

\usepackage[capitalize]{cleveref}
\usepackage[utf8]{inputenc}
\usepackage[T1]{fontenc}
\usepackage{mathtools}
\crefname{section}{Sec.}{Secs.}
\Crefname{section}{Section}{Sections}
\Crefname{table}{Table}{Tables}
\crefname{table}{Tab.}{Tabs.}

\crefname{appendix}{Sec.}{Secs.}
\Crefname{appendix}{Section}{Sections}

\def\wacvPaperID{78} % *** Enter the WACV Paper ID here
\def\confName{WACV}
\def\confYear{2025}

\begin{document}

\title{MFTIQ: Multi-Flow Tracker with Independent Matching Quality Estimation}

% \and
\author{Jonas Serych, Michal Neoral, Jiri Matas\\
  {\small CMP Visual Recognition Group, Faculty of Electrical Engineering, Czech Technical University in Prague} \\
  {\tt\small \{serycjon,neoramic,matas\}@fel.cvut.cz}
}
\maketitle

%%%%%%%%% ABSTRACT
\begin{abstract}
   In this work, we present MFTIQ, a novel dense long-term tracking model that advances the Multi-Flow Tracker (MFT) framework to address challenges in point-level visual tracking in video sequences.
   MFTIQ builds upon the flow-chaining concepts of MFT, integrating an Independent Quality (IQ) module that separates
%   occlusion and
   correspondence quality estimation from optical flow computations.
   This decoupling significantly enhances the accuracy and flexibility of the tracking process, allowing MFTIQ to maintain reliable trajectory predictions even in scenarios of prolonged occlusions and complex dynamics.
   Designed to be ``plug-and-play'', MFTIQ can be employed with any off-the-shelf optical flow method without the need for fine-tuning or architectural modifications.
   Experimental validations on the TAP-Vid Davis dataset show that MFTIQ with RoMa~\cite{edstedt2023roma} optical flow not only surpasses MFT but also performs comparably to state-of-the-art trackers while having substantially faster processing speed.
   Code and models available at \url{https://github.com/serycjon/MFTIQ} .
\end{abstract}

%%%%%%%%% BODY TEXT

% =====================================================================================
% ==                                                                                 ==
% ==                                  INTRODUCTION                                   ==
% ==                                                                                 ==
% =====================================================================================
\section{Introduction}
\label{sec:intro}
Point-level visual tracking is a hot research topic~\cite{li2024taptr,karaev2023cotracker,doersch2023tapir,doersch2022tap}.
Instead of the classical task of tracking objects by bounding boxes~\cite{Bolme2010, Kalal2011, danelljan2018atom} or segmentation masks\cite{kristan2020eighth, perazzi2016davis}, the goal is to track arbitrary points lying on surfaces in the scene.
The resulting point correspondences are useful for various downstream applications, like SLAM\cite{Geiger2012, Nair2020} or motion prediction\cite{Wu2020}.
While most current methods\cite{doersch2022tap, doersch2023tapir, doersch2024bootstap, Xiao2024, karaev2023cotracker} focus on \emph{sparse} point-tracking, applications like 3D reconstruction, video editing, or augmented reality, benefit from \emph{dense} correspondences, \ie, correspondences estimated for every pixel of the initial video frame.

Traditionally, long-range dense tracking may be achieved by sequential chaining of optical flow (OF).
However this approach has major drawbacks.
The estimated trajectory drift over time due to error accumulation, and tracking stops to be reliable in presence of occlusion since the sequential chaining has no mechanism to recover.
\begin{figure}
  \centering
  \begin{overpic}[width=0.35\textwidth]{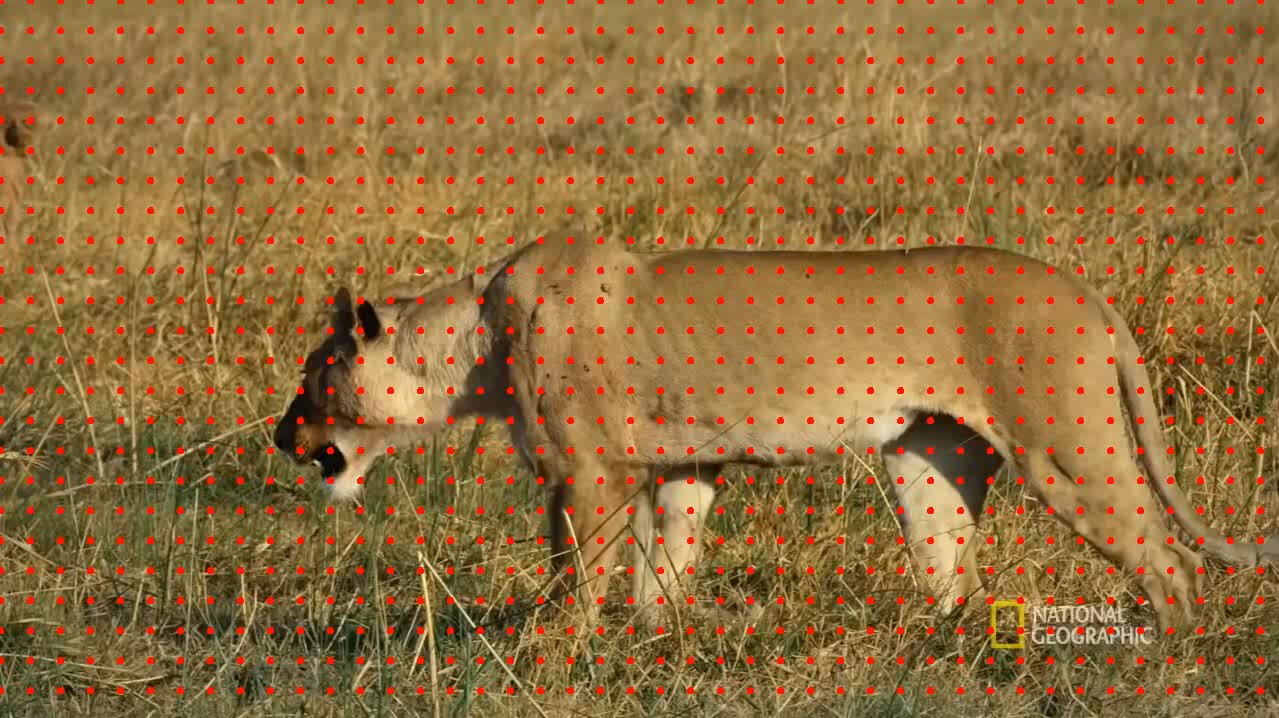}
    \put (2,50) {\color{white} \textsf{\textbf{Input - frame 1}}}%$\Img_1$}}}
  \end{overpic}
  \begin{overpic}[width=0.35\textwidth]{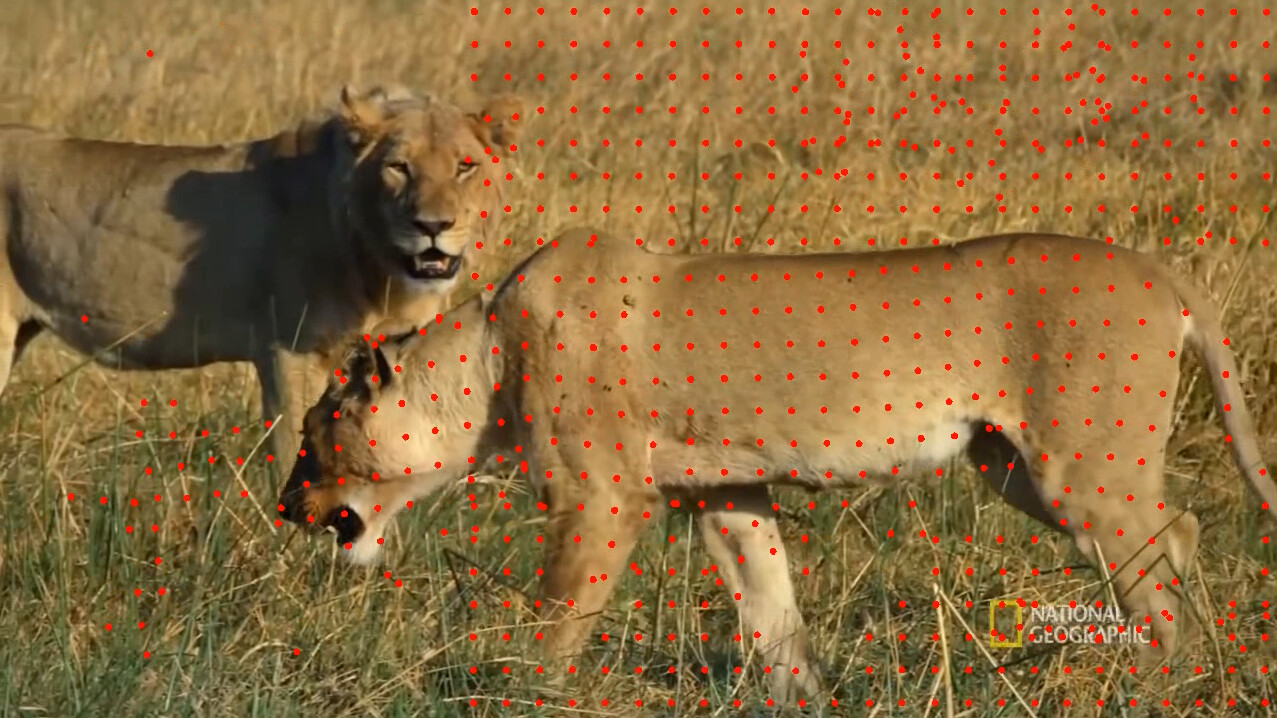}
    \put (2,50) {\color{white} \textsf{\textbf{MFT - frame 80 }}}%$\Img_{80}$}}}
  \end{overpic}
  \begin{overpic}[width=0.35\textwidth]{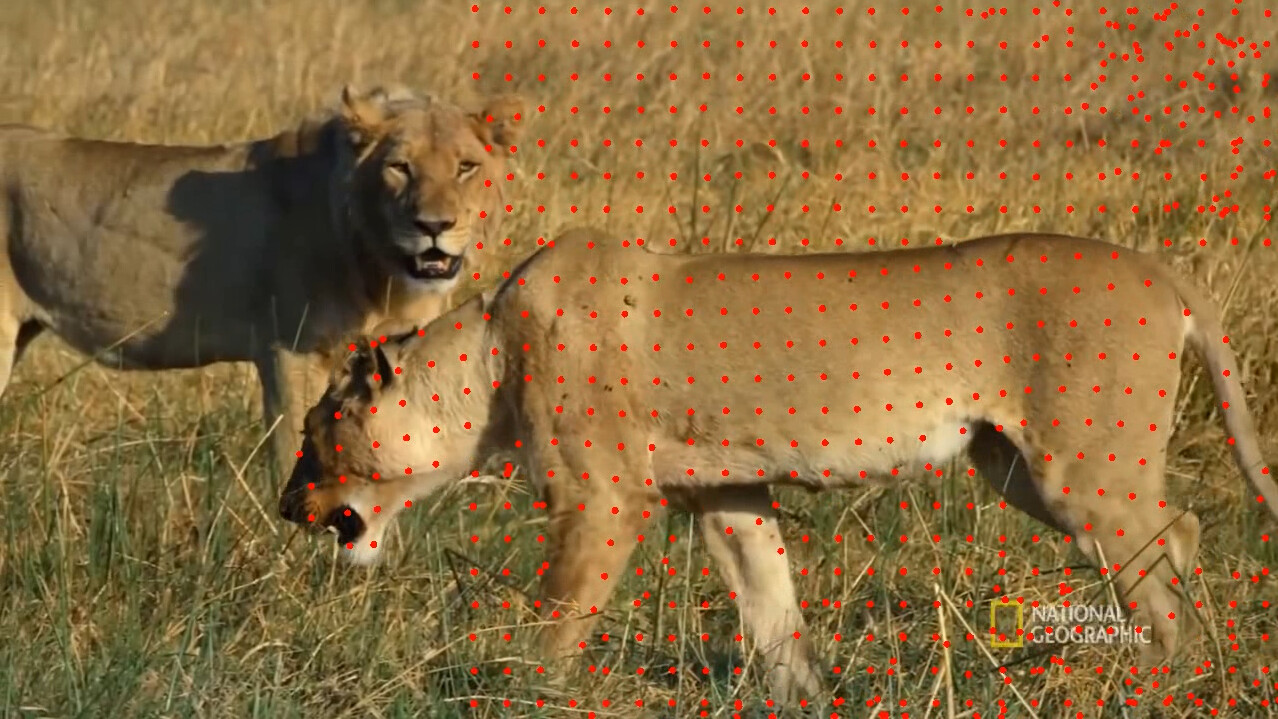}
    \put (2,50) {\color{white} \textsf{\textbf{MFTIQ - frame 80}}}%$\Img_{80}$}}}
  \end{overpic}
  \caption{\textbf{ Dense long-term tracking -- \method{MFT} and \method{MFTIQ} comparison}.
  Visualisation of query positions \textit{(red)} tracked from frame 1 to frame 80.
  \method{MFTIQ} generates a lower number of false re-detections than \method{MFT}, especially on the grass in bottom-left, which was out-of-view on frame 1.
  Best viewed zoomed-in and in color.
  }
  \label{fig:introfig}
\end{figure}
Recently, Multi-Flow Tracker (\method{MFT})~\cite{neoral2024mft} revisited flow chaining~\cite{crivelli2012optical,conze2016multi} for not only
consecutive frames, but also for temporally distant frame pairs.
\method{MFT} produces long, dense trajectories by selecting the most reliable chain of optical flows for each tracked point.
The flow chain reliability is determined by accumulating uncertainties and occlusion state computed for each optical flow in the flow chain.
However, this uncertainty accumulation can lead to error accumulation and drift.
Moreover, \method{MFT} is tightly coupled with the \method{RAFT} optical flow, but it has been shown~\cite{jelinek2024dense} that other OF methods perform better.

We propose \method{MFTIQ}, a dense, long-term tracker.
Like other dense point-trackers~\cite{cho2024flowtrack,moing2023dense,neoral2024mft,wang2023omnimotion,jelinek2024dense} it is based on optical flow computation.
We design the method to be trained once and then work with an arbitrary optical flow method in a ``plug-and-play'' fashion and without any fine-tuning or architecture changes.
This allows the user to choose a suitable speed/performance trade-off by using the appropriate flow.
The \method{MFTIQ} generalizes to multiple optical flow methods not seen during training, as we have experimentally evaluated.
We expect future faster and/or higher quality flows to improve the proposed tracker performance for free, \ie without any re-training needed.

The proposed method replaces the flow-chain-selection method proposed in \method{MFT}, with an improved Independent Quality and occlusion estimation module (IQ).
Unlike the MFT method, which estimates occlusion and correspondence quality (uncertainty) jointly with optical flow estimation, MFTIQ decouples these estimations from the optical flow computation.
This separation also allows the occlusion and correspondence quality to be estimated directly for the optical flow chain between the \template and the current frame, without need for error-prone uncertainty accumulation.
\Cref{fig:introfig} shows an example where the \method{MFTIQ} strategy produces significantly less false re-detections than \method{MFT}.

MFTIQ achieves results comparable to state-of-the-art trackers when using \method{RoMa}~\cite{edstedt2023roma} as the optical flow estimator and consistently outperforms MFT across most tested optical flow methods.
It is important to note that \method{MFTIQ} was not trained with \method{RoMa}, highlighting the ``plug-and-play'' functionality of the proposed method.
Moreover, for dense tracking, MFTIQ is significantly faster than state-of-the-art trackers, even with the slowest optical flow methods tested.
\method{MFTIQ} is also causal, \ie, it only uses the current frame and the previous ones, which is not the case for most point-trackers.

In this work, we introduce \method{MFTIQ}, a novel dense, long-term tracking method improving on the MFT~\cite{neoral2024mft} flow-chaining idea.
\textbf{Our contributions are as follows}:
1) We have developed the Independent Quality (IQ) module, which decouples occlusion and correspondence quality estimation from optical flow computation.
This separation enhances tracking accuracy and flexibility.
2) MFTIQ features ``plug-and-play'' functionality, allowing integration with any off-the-shelf optical flow method without re-training or fine-tuning.
This flexibility enables users to tailor tracker performance to specific needs.
3) We conducted experimental evaluations using multiple optical flow methods, demonstrating that MFTIQ matches the performance of state-of-the-art trackers while being significantly faster in dense tracking scenarios.

% =====================================================================================
% ==                                                                                 ==
% ==                                  RELATED WORK                                   ==
% ==                                                                                 ==
% =====================================================================================
\section{Related Work}
\label{sec:related-work}

\paragraph{Optical flow}
is a fundamental problem~\cite{Horn1981} in computer vision in which the pixel-level displacement between pair of frames is to be densely estimated.
Most of the current methods are based on learning~\cite{Dosovitskiy2015, Ilg2017, shi2023flowformerpp, Sun2018c, neoral2018continual, edstedt2023roma, zhang2024neuflow, dong2024memflow, Shi2023a}.
FlowNet~\cite{Dosovitskiy2015} introduced correlation cost-volume (CCV) to learning based optical flow estimation to provide similarity measurement between neighboring features from consecutive frames.
Later, RAFT~\cite{teed2020raft} employs 4D CCV for all pairs of pixel on lower resolution and iteratively estimates optical flow.
\method{FlowFormer}~\cite{Huang2022, shi2023flowformerpp} updates RAFT with transformer blocks.
\method{RoMa}~\cite{edstedt2023roma} is a dense wide-baseline stereo matcher that can be, however, used as optical flow estimator.
Both, \method{FlowFormer} and \method{RoMa} bring higher accuracy for the cost of slower processing time and bigger memory requirements.
This is addressed with \method{NeuFlow}~\cite{zhang2024neuflow} or \method{SEA-RAFT}~\cite{wang2024sea}, which focus more on efficient, higher speed computation.

While these methods are state-of-the-art in field of optical flow estimation, their possibility for employment in \emph{long-term} tracking are rather limited.
There are multi-frame optical flow approaches, such as \method{VideoFlow~\cite{Shi2023a}} or \method{MemFlow}~\cite{dong2024memflow}, but they are still focused on estimation of optical flow between adjacent frames rather than long-term optical flow and mentioned limitation of optical flow chaining remains.

\paragraph{Sparse long-term point-tracking} focus on tracking small number of query points on the object surface throughout the video.
Particle-video~\cite{sand2008particle} tracks only visible query points and fails to track continuously through occlusions, instead starting new tracks.
Revisiting this, \method{PIPs}~\cite{Harley2022} takes frames from fixed-size temporal window (8 frames) and estimate sparse point tracking with iterative updates.
They propose a strategy for linking the eight-frame tracks over longer period of time, however their method cannot recover from longer occlusions.
\method{TAP-Net}~\cite{doersch2022tap} computes CCV for each query point with each frame of a video and from it estimate occlusion and position by their two branch network.
\method{TAPIR}~\cite{doersch2023tapir} combines per-frame global-matching prediction of \method{TAP-Net} with refining process inspired by \method{PIPs}.
Current state-of-the-art -- \method{BootsTAP}~\cite{doersch2024bootstap} is the \method{TAPIR} tracker fine-tuned in a self-supervised fashion on large amount of in-the-wild videos.
The long training using 256 A100 GPUs on the 15M YouTube video clips is however too costly to reproduce for most researchers.

\method{CoTracker}~\cite{karaev2023cotracker} processes query points with a sliding-window transformer that enables multiple tracks to influence each other.
However, the best performance is achieved by tracking single query point at time, supplemented with auxiliary grid of queries.
\method{SpacialTracker}~\cite{Xiao2024} builds upon \method{CoTracker}.
Instead of tracking points in 2D, it lifts them to 3D using an off-the-shelf monocular depth estimation method.
All the mentioned methods can track densely by tracking the points one-by-one or in batches, but the resulting speed is low.

\paragraph{Dense long-term tracking}
approaches track all points from reference frame to current frame simultaneously.
\method{OmniMotion}~\cite{wang2023omnimotion} employs \method{NeRF}~\cite{Mildenhall2021} for modeling of a dynamic scene, enabling it to produce dense tracking outputs.
However, it relies on optical flow estimated between all pairs of frames followed by computationally demanding test-time training for each video sequence, making it impractical for general use.

DOT~\cite{LeMoing2024} estimates dense-point tracking by a two-stage process.
First, sparse point-tracks are estimated by \method{CoTracker}~\cite{karaev2023cotracker}.
Then they are densified and serve as an initialization for RAFT~\cite{teed2020raft} optical flow, which provides the final dense predictions.
\method{FlowTrack}~\cite{cho2024flowtrack} chains optical flow and corrects it by error compensation module utilizing optical flow forward-backward cycle consistency.

Recently, MFT~\cite{neoral2024mft} extends optical flow chaining by not only tracking between consecutive frames but also between frames that are temporally distant.
The chaining approach over various intervals between frames was addressed before~\cite{crivelli2012optical,conze2016multi}, however MFT proposed effective strategy where a long-dense trajectory is computed by evaluating the quality of track estimates among various combinations of chained flows, enabling it to maintain accurate tracking over longer sequences than typical flow-based methods.
\method{MFT-RoMa}~\cite{jelinek2024dense} integrates wide-baseline dense matchers \method{DKM}~\cite{edstedt2023dkm} and \method{RoMa}~\cite{edstedt2023roma} into MFT, further increasing its performance.

% =====================================================================================
% ==                                                                                 ==
% ==                                     METHOD                                      ==
% ==                                                                                 ==
% =====================================================================================
\section{Method}
\label{sec:method}

\begin{figure}
  \centering
  \includegraphics[page=4,width=0.5\textwidth]{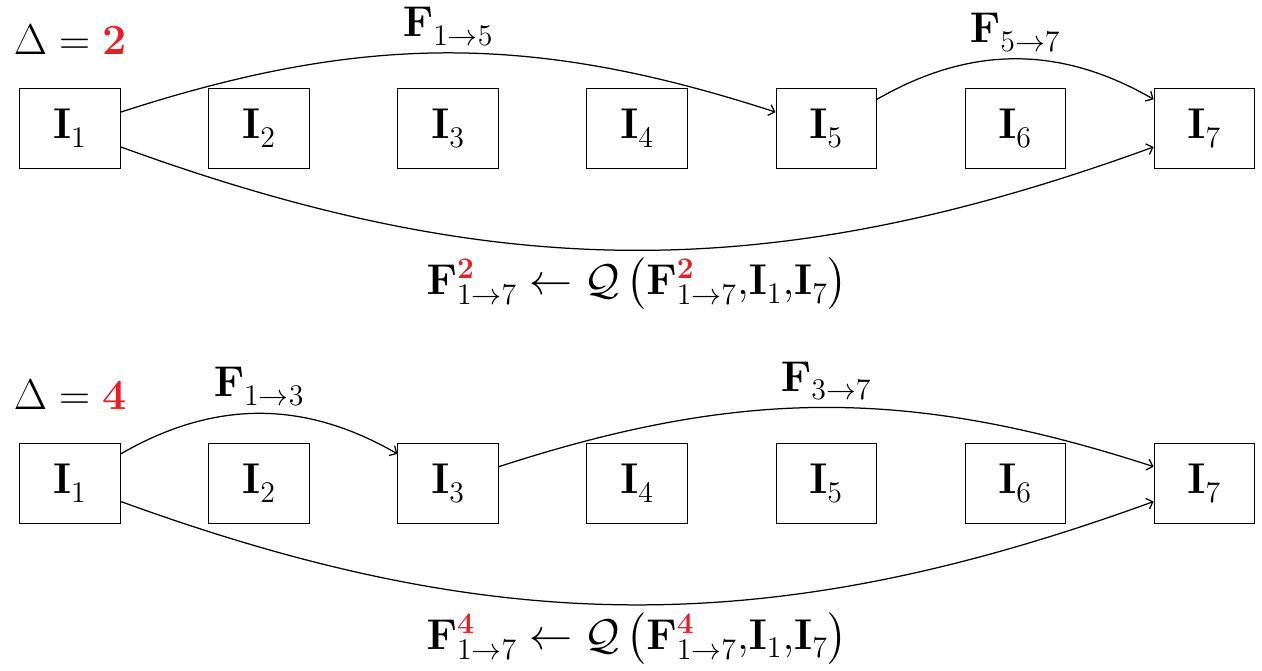}
  \caption{\textbf{Example of MFTIQ optical flow chaining strategy} for estimating the flow between $I_1$ and the current frame $I_{t=7}$.
    \method{MFTIQ} constructs a flow chain $\Flow_{\ft{1}{7}}^{\Delta}$ by going through an \emph{intermediate frame} $\Img_{7-\Delta}$.
    This is done for multiple values of $\Delta$, here shown for $\Delta = \mathbf{\color{red}2}$ and $\Delta = \mathbf{\color{red}4}$.
    The most reliable flow chain $\Flow_{\ft{1}{7}}^{\Delta^{\star}}$ is selected independently in each pixel based on flow quality $\Qualitymethod \left( \Flow_{\ft{1}{7}}^{\Delta}, \Img_1, \Img_7 \right)$ assigned to each chain $\Flow_{\ft{1}{7}}^{\Delta}$ by a neural network, which takes the flow chain, the \template frame, and the current frame as inputs.
    Note that the flow from the \template into the intermediate frame is itself a previously computed flow chain, while the flow from the intermediate frame into the current frame is output of an OF method.
  }
  \label{fig:mftiq-vs-mft}
\end{figure}

\begin{figure}
\centering
\includegraphics[page=3,width=0.5\textwidth]{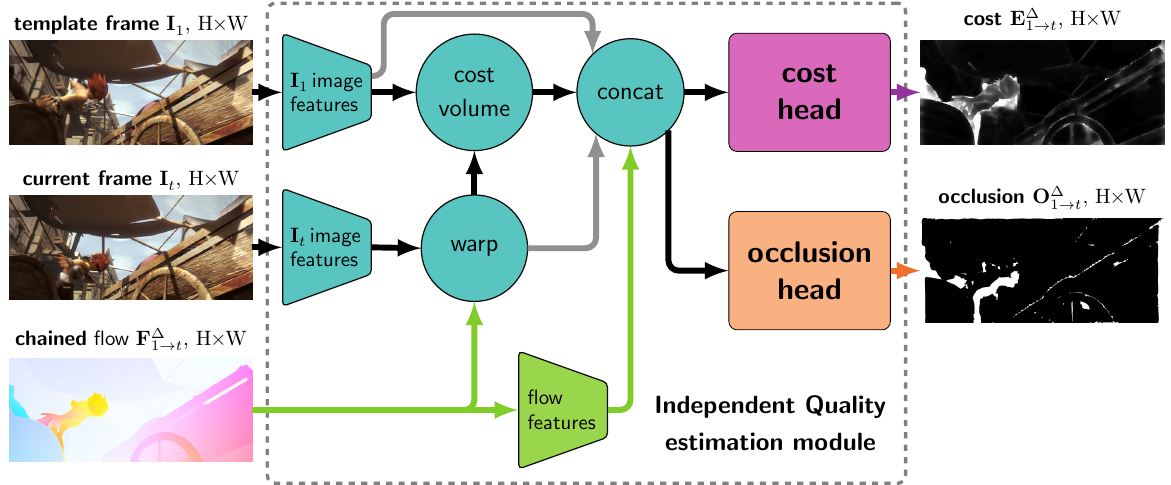}
\caption{\textbf{Overview of the Independent Quality (IQ) estimation network.}
  First, image features are extracted from the \template frame $\Img_1$ and the current frame $\Img_t$.
  Then, the current frame features are warped using the positions given by the chained flow $\Flow_{\ft{1}{t}}^{\Delta}$.
  The now-aligned feature maps are compared with a local (displacement up to $\pm3$) correlation cost-volume.
  Finally a concatenation of the features extracted from both images and the flow are concatenated with the cost-volume and processed by two small CNNs to output the occlusion map and the \costname map which together represent the quality of the input flow chain.
}
\label{fig:mftiq_scheme}
\end{figure}

We propose a dense long-term tracker based on chaining of optical flows computed from neighboring, but also from more distant frames.
We will first explain the flow-chaining technique that was used before~\cite{crivelli2012optical,conze2016multi} and recently revisited in \method{MFT}~\cite{neoral2024mft}.
Then we describe the proposed \method{MFTIQ} tracker and how it differs from \method{MFT}.

\paragraph{Task and notation.}
The task is to track all points from an initial frame to the rest of the video.
In particular, given a sequence of video frames $\left( \Img_t \right)_{t=1}^{T}$ with $H\stimes W$ resolution, the \method{MFTIQ} tracker computes \emph{long-term} flow fields $\left( \mat{F}_\ft{1}{t} \right)_{t=1}^T$ between the initial and the current frame.
For a 2D position $\vect{p}_A = (x, y)$ in image $\Img_A$, a flow field $\mat{F}_{\ft{A}{B}}$ bilinearly sampled at $\vect{p}_A$ gives a position in the image $\Img_B$ as $\vect{p}_{B} = \vect{p}_{A} + \sampleat{\mat{F}_{\ft{A}{B}}}{\vect{p}_A}$, where $\sampleat{\cdot}{\cdot}$ is the bilinear sampling.
\method{MFTIQ} also outputs binary segmentation maps $\left( \mat{O}_t \right)_{t=1}^T$ indicating that a point $\vect{p}_1$ is occluded or out-of-view in frame $\Img_t$ when $\sampleat{\mat{O}_{t}}{\vect{p}_1} > 0.5 .$

Both the flows $\Flow_{\ft{1}{t}},$ and the occlusion masks $\Occl_t$ have the full $H\stimes W$ resolution.
To simplify the explanation, we refer to the first frame $\Img_1$ of the video as \textit{\template frame}, $\Img_t$ as \emph{current frame}.

% =====================================================================================
% ==                                        MFT                                      ==
% =====================================================================================
\subsection{Optical Flow Chaining}
\label{subsec:chaining}
On each frame the proposed \method{MFTIQ} constructs flow fields $\Flow_{\ft{1}{t}}^{\Delta}$ as a \emph{chain} of flow field $\Flow_{\ft{1}{(t-\Delta)}}$ computed previously in an intermediate frame number $t-\Delta$ and a flow field from the intermediate to the current frame $\Flow_{\ft{(t-\Delta)}{t}}$.
The chaining operation samples the second flow field at the positions given by the first one such that
\begin{equation}
  \label{eq:flow-chain}
  \sampleat{\Flow_{\ft{1}{t}}^{\Delta}}{\vect{p}_1} = \sampleat{\Flow_{\ft{(t-\Delta)}{t}}}{\vect{p}_{1} + \sampleat{\Flow_{\ft{1}{(t-\Delta)}}}{\vect{p}_1}}
\end{equation}
Note that this can form an arbitrarily long chain of flows, as the $\Flow_{\ft{1}{(t-\Delta)}}$ was itself formed as a chain of flows.

Like in~\method{MFT}, the proposed \method{MFTIQ} computes a small number of such flows for varying logarithmically spaced time deltas $\Delta \in \Deltaset = \left\{1, 2, 4, 8, 16, 32\right\} \cup \{t - 1\}$, where the $\{t - 1\}$ stands for direct match, in which the optical flow is computed in single step between the \template and the current frame.
Flows for $\Delta \geq t$ are not computed.
This results in a set of candidate optical flow chains (each represented by a single flow field), from which the most reliable one is selected according to its \emph{quality}.

In the original \method{MFT}~\cite{neoral2024mft}, the quality was measured by flow uncertainty and occlusion.
In particular, the \method{RAFT}~\cite{teed2020raft} optical flow network was extended by two additional CNN heads, estimating the occlusion state and positional uncertainty in each pixel.
These two quantities were aggregated during the chaining of the flow fields to produce the overall positional uncertainty $\mat{U}_t^{\Delta}$ and occlusion state $\mat{O}_t^{\Delta}$ of the whole flow chain.
Finally these were used to select the most reliable flow chain $\Flow_{\ft{1}{t}}^{\Delta^{\star}}$ \emph{per-pixel} as
\begin{equation}
  \label{eq:mft-delta-selection}
  \sampleat{\Delta^{\star}}{\vect{p}_1} = \arg\,\min_{\Delta \in \Deltaset} \sampleat{\mat{U}_t^{\Delta}}{\vect{p}_1} + \infty \cdot \sampleat{\mat{O}_t^{\Delta}}{\vect{p}_1},
\end{equation}
where multiplying the occlusion by infinity ensures that a flow chain that was occluded at any time is only selected if there is no unoccluded, \ie, better chain.

The \method{MFT} approach has three major drawbacks.
First, the uncertainty estimation and the chaining of the uncertainty scores need to be well calibrated not to be overly optimistic or pessimistic.
This is not the case in \method{MFT} which is slightly pessimistic, leading to a strong preference of flow chains with small number of links, \ie, a preference of $\Delta^{\star}$ being large.
In our experience, this often happens even though there are more accurate longer chains (more links with smaller $\Delta^{\star}$s) available.
On the other hand, when the uncertainty of a single incorrect chain link is optimistically low, the tracker drifts and tracks a different point from that moment on.
Finally, it is not straightforward to use different optical flow methods due to the direct integration of occlusion and uncertainty into the flow network architecture.

The proposed \method{MFTIQ}, addresses these issues with its \emph{Independent Quality} (IQ) module, which replaces the problematic chaining of uncertainties with a direct estimation of the quality of the chained optical flow.
The best delta is again selected per-pixel similarly to~\cref{eq:mft-delta-selection} as
\begin{equation}
  \label{eq:mftiq-delta-selection}
  \sampleat{\Delta^{\star}}{\vect{p}_1} = \arg\,\min_{\Delta \in \Deltaset} \sampleat{\Cost_t^{\Delta}}{\vect{p}_1} + M \cdot \sampleat{\mat{O}_t^{\Delta}}{\vect{p}_1},
\end{equation}
where $M$ is a large constant used instead of the $\infty$ in~\cref{eq:mftiq-delta-selection}.
This still ensures that unoccluded chains are always preferred, but also preserves the ordering by $\sampleat{\mat{E}_t^{\Delta}}{\vect{p}_1}$ when all the candidate flow chains contain an occlusion.
The \costname map $\Cost$ functions analogously to the \method{MFT} flow chain uncertainty $\mat{U}$, but is trained with a different cost function.
\Costname $\Cost$ is analogous to MFT uncertainty $\Unc$ in that the lower values means higher positional accuracy.

Most importantly, we propose to estimate the \costname $\Cost$ and the occlusion map $\Occl$ directly as a function of the chained flow $\Flow_{\ft{1}{t}}^{\Delta}$ and the two images it relates to, $\Img_1$ and $\Img_t$.
\begin{equation}
  \label{eq:quality}
  \{\Cost_{t}^{\Delta}, \Occl_{t}^{\Delta}\} = \Qualitymethod\left( \Flow_{\ft{1}{t}}^{\Delta}, \Img_1, \Img_t  \right).
\end{equation}
The independent quality estimation function $\Qualitymethod$ is implemented as a neural network.
It takes the chained optical flow, the \template frame, and the current frame as inputs and estimates the \costname map $\Cost \in \mathbb{R}_{0+}^{H\times W}$ and the occlusion map $\Occl \in \{0, 1\}^{H\times W}$.
An example diagram of the \method{MFTIQ} flow chaining and selection is shown in~\cref{fig:mftiq-vs-mft}.

% =====================================================================================
% ==                           Occlusion and Quality Module                          ==
% =====================================================================================
\subsection{Flow Quality Estimation}
\label{subsec:quality}
In this section, we detail the architecture and the training of the proposed quality estimation network $\Qualitymethod$, shown in overview in~\cref{fig:mftiq_scheme}.
First, we extract image features to produce a $\frac{H}{4}\times\frac{W}{4}$ feature map.
In particular, both $\Img_1$ and $\Img_t$ are processed by the \method{DinoV2}~\cite{Oquab2023} network.
We bilinearly upscale the resulting coarse $\frac{H}{14}\times \frac{W}{14}$ feature map into the target $\frac{H}{4}\times\frac{W}{4}$ resolution.
To add more spatially fine-grained information, we also compute the \method{ImageNet1K}-pre-trained \method{ResNet50}~\cite{he15resnet} CNN features and features from a shallow CNN.
See~\cref{sec:image-feat-extraction} in supplementary for more details.
We resize all the resulting feature maps into the $\frac{1}{4}$ resolution and compress them with a convolutional layer to have 32 channels each.
\paragraph{Warping + Cost Volume}
The next stage in our process is the formation of a local Correlation Cost Volume (CCV), which serves to measure the similarity between the corresponding (as predicted by the optical flow chain) features, while also considering adjacent pixel information.
To perform this, the feature maps from the current frame $I_t$, are warped to the \template frame $I_1$ using the chained optical flow $\Flow_{\ft{1}{t}}$, which is scaled to match the featuremap resolution.
Then a local CCV (maximum displacement of $3$px on the featuremap resolution) is independently computed for each input feature map, like in \method{FlowNet}~\cite{Dosovitskiy2015}.

Finally, we concatenate the feature similarities computed by the cost-volumes with the image features and features computed from the chained optical flow, more details in supplementary~\cref{sec:feat-aggr-after}.
The resulting $\frac{H}{4} \times \frac{W}{4}$ featuremap with 422 channels is then used to estimate the flow-chain quality.

\paragraph{Flow-Chain Quality Estimation}
We use two three-layer CNN heads, each followed by a bilinear upsampling to the full image resolution, to estimate the \costname and occlusion maps.
The occlusion estimation CNN classifies each pixel as either occluded or non-occluded and is trained using standard binary cross-entropy loss, denoted as $\mathcal{L}^{\mathrm{occl}}$.

The \costname is constructed from $M = 5$ binary classifiers again trained by binary cross-entropy loss $\mathcal{L}^{\mathrm{match}\,\theta}$.
Pixels that have the flow end-point-error (EPE, euclidean distance from the ground-truth) over $\theta$px or are occluded belong to the positive class, while the visible and precisely matched (EPE under $\theta$px) belong to the negative class.
The binary classifiers differ in the EPE threshold $\theta \in \{1, 2, 3, 4, 5\}$, ranging from 1 to 5px.

During inference, the final \costname map is constructed as a weighted average of the soft (Sigmoid activation) classification maps $\Cost_{\theta}$, 
\begin{equation}
  \Cost = \sum\limits_{\theta = 1}^{M} 2^{\theta - 1} \Cost_{\theta}.
\end{equation}
The $\Cost$ should be low for well matched points and high for poorly matched or occluded points.

The overall training loss, $\mathcal{L}$, is computed as follows:
\begin{equation}
  \label{eq:total-loss}
  \mathcal{L} = \frac{1}{H \times W} \sum_{i=1}^{H \times W} \mathbf{V}_i \left( \mathcal{L}_i^{\mathrm{occl}} + \frac{1}{M}\sum\limits_{\theta=1}^M\mathcal{L}_{i}^{\mathrm{match}\,\theta} \right),
\end{equation}
where $\mat{V}_i$ is a binary ground-truth validity flag of pixel $i$.

\subsection{Implementation Details}
\label{subsec:implementation-details}
We trained the independent quality network using a synthetic dataset from the Kubric rendering tool~\cite{Greff2022}.
The dataset includes 200 sequences with a variable number of static and dynamic objects rendered at a $1024 \stimes 1024$ resolution, each 240 frames long.
The sequence length is much longer than the typically used 24 or 48 frames.
We had to ensure that the objects do not become static after falling to the ground as in the default Kubric scenario, otherwise the long sequences would not bring much. 
To do this and keep the objects non-intersecting, we left the default Kubric physical engine to simulate the scene for 48 frames, after which we disabled it and replayed the simulated motions back and forth for the rest of the video.
The camera motion is generated independently, with the panning from \method{TAPIR} and a random camera shake to introduce motion blur and make the camera movement more realistic.
Due to the independent non-looping motion of the camera, the resulting video is not repetitive and information-rich for the whole duration.

The training involved sampling random image pairs with temporal separations, \ie, the flow $\Delta$, ranging from 2 to 150 frames.
We generated a pre-sampled set of 20,000 training pairs with dense\footnote{The original version of the Kubric tool supports only sparse ground truth generation for point-tracking tasks.} ground truth optical flow, occlusion, and validity masks $\mathbf{V}$.
The input OFs were uniformly drawn from the ground truth flow, \method{RAFT}~\cite{teed2020raft}, and ground-truth-initialized \method{FlowFormer++}~\cite{shi2023flowformerpp}, computed directly between the two input images.
This generates plausible long-term input OFs without tracking the whole sub-sequence in each training step.

Both the optical flow and the input images were augmented and resized to $368 \times 768$ pixels.

The training was conducted on a single RTX A5000 GPU for approximately one day using a batch size of 8 for 200,000 iterations, with an initial learning rate of $2.5 \times 10^{-3}$ and OneCycleLR~\cite{Smith2019} learning rate policy.

We set $\Delta \in \left\{ 1, 2, 4, 8, 16, 32, t-1 \right\}$, \ie, the same as in \method{MFT}.
See~\cref{sec:delta-value-ablation} in supplementary for experimental evaluation of different $\Delta$-set configurations.

\paragraph{Inference-time caching}
To speed up the proposed \method{MFTIQ} tracker, we cache and re-use intermediate results where possible.
Namely, the image features are needed multiple times per frame and especially the \method{DinoV2} network is slow, so we cache them in GPU memory.
We also cache the optical flows, which is useful when tracking from multiple query frames, like in the strided \dataset{TAP-Vid}.
If the application allows it, both the image features and the optical flows can be precomputed to get fast tracking.
Timing details are reported in supplementary~\cref{sec:timing}.

% =====================================================================================
% ==                                                                                 ==
% ==                                   EXPERIMENTS                                   ==
% ==                                                                                 ==
% =====================================================================================
\section{Experiments}
\label{sec:experiments}
Since there is no \emph{dense} long-term tracking benchmark, we evaluate the proposed \method{MFTIQ} tracker on standard \emph{sparse} point-tracking datasets \dataset{TAP-Vid}~\cite{doersch2022tap} and \dataset{RoboTAP}~\cite{vecerik2023robotap}.
We also evaluate on the \dataset{POT-210}~\cite{liang2017planar} dataset for planar object tracking, which contains challenging scenarios different from \dataset{TAP-Vid} and was not leveraged for point-tracking evaluation before.

\subsection{Point-Tracking Benchmark}
\label{sec:point-track-benchm}
The point-tracking is typically evaluated using three metrics introduced in \dataset{TAP-Vid}~\cite{doersch2022tap}.
The \deltaavg measures the percentage of cases where the euclidean distance between the predicted and the ground-truth position is smaller than a threshold, averaged over five thresholds of 1, 2, 4, 8, and 16 pixels.
This evaluation is done on coordinates re-scaled to $256 \times 256$ resolution.
The quality of occlusion prediction -- occlusion accuracy (OA) -- is measured by standard binary classification accuracy.
Finally, the average Jaccard (AJ) metric combines the position and occlusion accuracy into an unified score.
Please refer to~\cite{doersch2022tap} for details.

There are two evaluation modes, \emph{first} and \emph{strided}.
In the \emph{first} mode, trackers are initialized on the first frame where the particular point is visible and left to track until the end of the video.
In the \emph{strided} mode, every fifth frame is taken as an initialization frame.
Trackers are initialized on all annotated points visible in the particular frame and left to track in both directions until the start and until the end of the video.
The resulting tracks are shorter (half the video length on average), making the task simpler.
Also, in the \emph{first} mode, the query points are often on the object boundary or just after de-occlusion, further complicating the tracking.

The \dataset{TAP-Vid DAVIS}~\cite{doersch2022tap} dataset contains 30 videos from~\cite{pont-tuset17davis}, mostly containing people and animals, with one or a few salient objects moving against a background.
The \dataset{TAP-Vid Kinetics}~\cite{doersch2022tap} has 1189 videos from Kinetics-700~\cite{carreira2019short,carreira2017quo} human action recognition dataset.
The \dataset{RoboTAP}~\cite{vecerik2023robotap} dataset contains 265 videos of robotic arms picking up and dropping objects in a lab scenario.
All the datasets have point tracks semi-automatically annotated for around 20 points in each video, including the visibility state.

\paragraph{Plug-n-Play Optical Flow}
\begin{table}
  \setlength{\tabcolsep}{5pt}
  \centering
  \begin{adjustbox}{max width=0.48\textwidth}
  \begin{tabular}{lcccrr}
  \toprule
                                                                &               &                     &               & \multicolumn{2}{c}{OF runtime [ms] $\downarrow$} \\
      method                                                    & AJ $\uparrow$ & \deltaavg$\uparrow$ & OA $\uparrow$ & 512x512  & 720x1080                              \\
    \midrule
    \method{MFT}~\cite{neoral2024mft}                           & 56.28         & 71.03               & 86.96         & 47       & 142                                   \\
    \midrule
    \method{MFTIQ} with                                         &               &                     &               &          &                                       \\
%    \midrule
    \hspace{0.5em}\method{RAFT}~\cite{teed2020raft}             & 60.54         & 74.22               & 84.42         & 47       & 142                                   \\
    \hspace{0.5em}\method{GMFlow}~\cite{Xu2022b}                & 55.28         & 69.83               & 83.55         & 24       & 137                                   \\
    \hspace{0.5em}\method{NeuFlow}~\cite{zhang2024neuflow}    & 55.73         & 70.26               & 80.87         & 10       & 18                                    \\
    \hspace{0.5em}\method{GMFlow-R}~\cite{Xu2022b}            & 59.57         & 73.38               & 86.49         & 69       & 335                                   \\
    \hspace{0.5em}\method{NeuFlowV2}~\cite{Zhang2024a}          & 56.92         & 70.97               & 81.59         & 7        & 8                                     \\
    \hspace{0.5em}\method{RAPIDFlow}~\cite{Morimitsu2024}       & 59.56         & 73.14               & 84.37         & 32       & 55                                    \\
    \hspace{0.5em}\method{LLA-Flow}~\cite{Xu2023}             & 61.78         & 75.18               & 85.44         & 117       & 475                                    \\
    \hspace{0.5em}\method{MemFlow}~\cite{dong2024memflow}       & 62.30         & 75.97               & 85.95         & 121      & 610                                   \\
    \hspace{0.5em}\method{FFormer++}~\cite{shi2023flowformerpp} & 62.72         & 76.22               & 86.34         & 142      & 782                                   \\
    \hspace{0.5em}\method{RPKNet}~\cite{Morimitsu2024a}       & 62.78         & 76.61               & 86.39         & 126       & 174                                    \\
    \hspace{0.5em}\method{SEA-RAFT}~\cite{Wang2024b}            & 63.51         & 77.18               & 86.22         & 34       & 105                                   \\
    \hspace{0.5em}\method{RoMa}~\cite{edstedt2023roma}          & \frst{65.67}  & \frst{79.82}        & \frst{87.75}  & 714      & 729                                   \\
    % \rowcolor{blue!10}%
  \bottomrule
  \end{tabular}
  \end{adjustbox}
  \caption{\dataset{TAP-Vid DAVIS}~\cite{doersch2022tap} (strided) evaluation with single \method{MFTIQ} model using various OF methods.
    The first two rows compare the original \method{MFT} with the proposed \method{MFTIQ} both using the \method{RAFT}~\cite{teed2020raft} OF.
    The rest of the table shows \method{MFTIQ} results when used with different OF methods.
    Runtime of a single OF computation shown on right.
  }
  \label{tab:MFTIQ_OF}
\end{table}
\begin{figure*}
  \centering
  \includegraphics[width=0.49\textwidth]{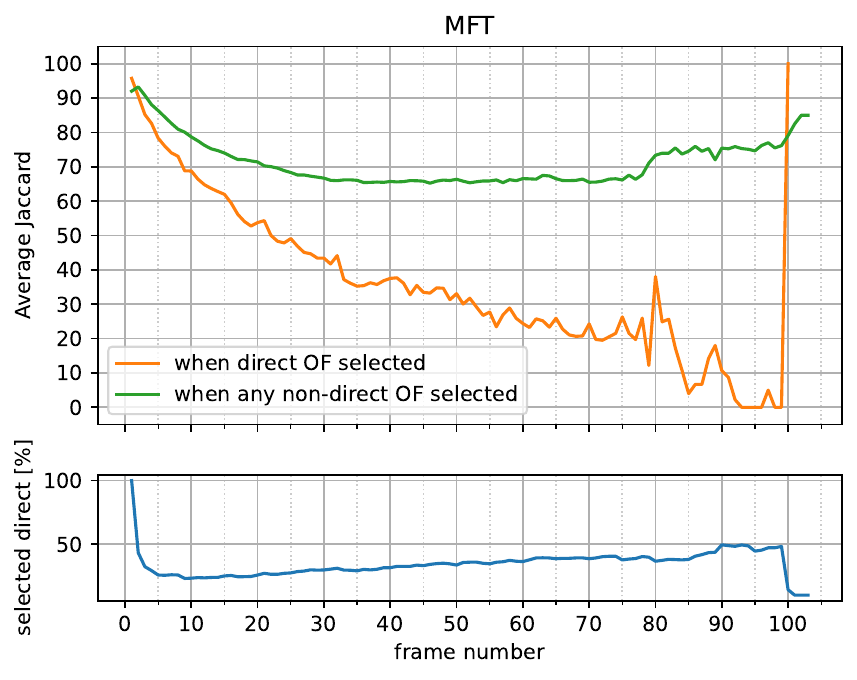}%
  \includegraphics[width=0.49\textwidth]{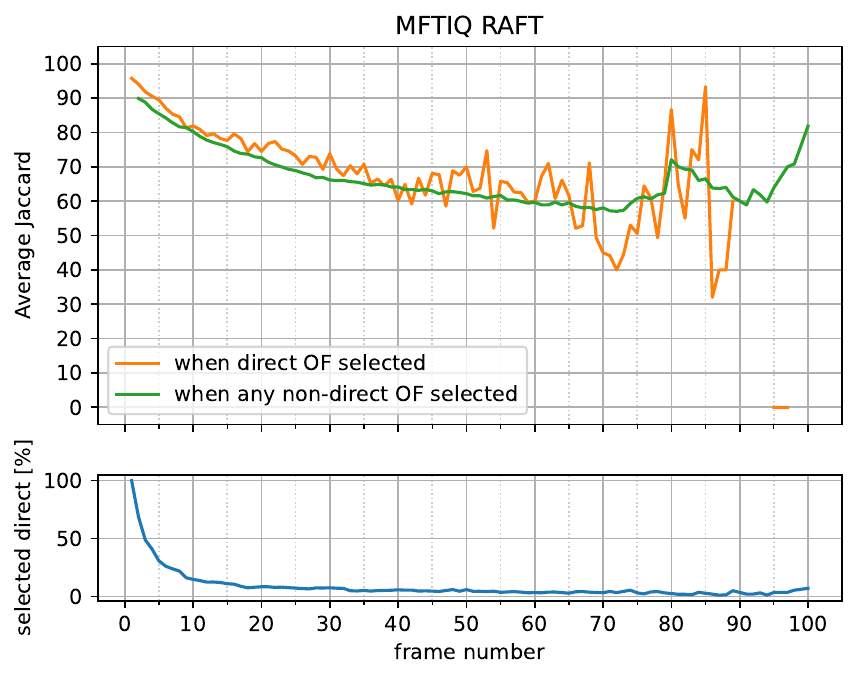}
  \caption{\textbf{Comparison of flow candidate selection in MFT \textit{(left)} and MFTIQ \textit{(right)}.}
    \method{MFT} often selects \textit{(bottom left)} the \emph{direct} optical flow, \ie the flow chain with $\Delta = t - 1$ with probability increasing during the video.
    The probability of the selected \emph{direct} flow to be accurate as measured by Average Jaccard (AJ) is, however, decreasing with time \textit{(orange)}.
    In contrast, the proposed \method{MFTIQ} \textit{(right)} chooses the direct optical flow more conservatively \textit{(bottom)} and mostly when it has high accuracy \textit{(orange)}.
    Both methods are evaluated on \method{TAP-Vid DAVIS} strided using \method{RAFT} OF.
    Non-direct OF accuracy \textit{(green)} represents the average over all cases, when some $\Delta \neq t-1$ was selected.
  }
  \label{fig:mft-vs-mftiq-chaining}
\end{figure*}
After training the \method{MFTIQ} flow quality estimation with \method{RAFT}~\cite{teed2020raft} and ground-truth initialized \method{FlowFormer++}~\cite{shi2023flowformerpp}, we fixed the model and evaluated it with various different OF methods.
The~\cref{tab:MFTIQ_OF} shows that the \method{RAFT}-based \method{MFTIQ} already outperforms the original \method{MFT}.
More importantly, we get even better results when using other off-the-shelf optical flows and dense matchers.
The best performance is achieved with the wide-baseline matcher \method{RoMa}~\cite{edstedt2023roma} thanks to its ability to match densely both between consecutive and between more distant frames.
\Cref{tab:MFTIQ_OF} also lists the runtime of the respective OF methods, measured on a RTX A5000 GPU\@.
While the best-performing \method{RoMa} is also the slowest on smaller images, it scales better than the second-best \method{FlowFormer++} to larger images.
Depending on the intended application, one could also use a fast optical flow method, such as \method{NeuFlow}~\cite{zhang2024neuflow}, for a cost of reduced tracking quality.
For the rest of the experiments we use the \method{RoMa}-based \method{MFTIQ}.

\paragraph{Main point-tracking results}
The overall results of the proposed \method{RoMa}-based \method{MFTIQ} tracker are shown in~\cref{tab:main-results}.
\method{MFTIQ} achieves the best (\dataset{DAVIS}) and the second-best (\dataset{RoboTAP}, \dataset{Kinetics}) position accuracy \deltaavg.
This is thanks to the quality of the used \method{RoMa} dense matcher.
Note that \method{RoMa} was used with the original \method{MFT} in \method{MFT-RoMa}~\cite{jelinek2024dense}, however due to better flow quality estimation the proposed \method{MFTIQ} performs much better on all metrics.
Also we have designed \method{MFTIQ} to be independent on the OF method, so we expect it to get better with future even-higher-quality optical flows and dense matchers without re-training.

The occlusion accuracy (OA) of \method{MFTIQ} is comparatively lower, also affecting the overall AJ score.
While it is an improvement over \method{MFT}, achieving state-of-the-art occlusion accuracy is yet an open challenge.

While \method{MFTIQ} does not achieve performance as good as the recent sparse point-trackers, it tracks densely and out-performs the original \method{MFT}.
Note that the point-trackers in~\ref{tab:main-results} are not \emph{causal}, \ie, the trackers can ``see'' into the future which is helpful to resolve occlusions.
Both \method{MFT} and \method{MFTIQ} only use the previous frames.
For dense tracking the inference time is significantly faster than methods with similar accuracy, as measured by the points-per-second metric in~\cref{tab:main-results}, more timing details in supplementary~\cref{sec:timing}.

\begin{table*}
  \centering
  \setlength{\tabcolsep}{5pt}
  \begin{tabular}{l|c|ccc|ccc|ccc|ccc}
    \toprule
                                                 &                 & \multicolumn{3}{c|}{\dataset{DAVIS} strided}      & \multicolumn{3}{c|}{\dataset{DAVIS} first}        & \multicolumn{3}{c|}{\dataset{RoboTAP} first}      & \multicolumn{3}{c}{\dataset{Kinetics} first}              \\
    method                                       & PPS$\uparrow$   & AJ$\uparrow$ & \deltaavg$\uparrow$ & OA$\uparrow$ & AJ$\uparrow$ & \deltaavg$\uparrow$ & OA$\uparrow$ & AJ$\uparrow$ & \deltaavg$\uparrow$ & OA$\uparrow$ & AJ$\uparrow$ & \deltaavg$\uparrow$ & OA$\uparrow$ \\
    \midrule
    \method{TAP-Net}~\cite{doersch2022tap}       &   $\dagger$ 555 & 38.4         & 53.1                & 82.3         & 33.0         & 48.6                & 78.8         & 45.1         & 62.1                & 82.9         & 38.5         & 54.4                & 80.6         \\
    \method{CoTracker}~\cite{karaev2023cotracker}&  $\ddagger$ 0.8 & 64.8         & 79.1                & 88.7  & \scnd{60.6}  & \scnd{75.4}         & \frst{89.3}  & 54.0         & 65.5                & 78.8         & 48.7         & 64.3                & \frst{86.5}  \\
    \method{TAPIR}~\cite{doersch2023tapir}       &   $\dagger$ 200 & 61.3         & 72.3                & 87.6         & 56.2         & 70.7                & 86.5         & 59.6         & 73.4                & \frst{87.0}  & \scnd{49.6}  & 64.2                & 85.0         \\
    \method{BootsTAP}~\cite{doersch2024bootstap} &            --   & \frst{66.4}  & 78.5         & \frst{90.7}  & \frst{61.4}  & 74.0                & 88.4  & \frst{64.9}  & \frst{80.1}         & \scnd{86.3}  & \frst{54.7}  & \frst{68.5}         & \scnd{86.3}  \\

    \method{DOT}~\cite{LeMoing2024} &            2473   & \scnd{65.9}  & \scnd{79.2}         & \scnd{90.2}  & 60.1  & 74.5                & \scnd{89.0}  & -  & -         & - & 48.4  & 63.8         & 85.2  \\
    \method{FlowTrack}~\cite{cho2024flowtrack} &           $*$ 499   & 63.2  & 76.3         & 89.2  & -  & -                & -  & -  & -         & -  & -  & -         & -  \\    

    \midrule
    \method{MFT}\cite{neoral2024mft}             &      10671      & 56.3         & 71.0                & 87.0         & 51.1         & 67.1                & 84.0         & --           & --                  & --           & 39.6         & 60.4                & 72.7         \\
    \method{MFT-RoMa}\cite{jelinek2024dense}     &             --  & 58.0         & 77.2                & 80.5         & 52.1         & 72.7                & 77.1         & --           & --                  & --           & --           & --                  & --           \\
    \textbf{MFTIQ (ours)}                        &        709      & 65.7  & \frst{79.8}         & 87.8         & 59.9         & \frst{75.5}         & 84.5         & \scnd{60.0}  & \scnd{77.5}         & 85.2         & 48.7         & \scnd{65.9}         & 85.2         \\
    \bottomrule
  \end{tabular}
  \caption{\method{MFTIQ RoMa} evaluation on \dataset{TAP-Vid}~\cite{doersch2022tap} and \dataset{RoboTAP}~\cite{vecerik2023robotap} benchmarks.
    On the \dataset{Kinetics} dataset, \method{MFTIQ} was evaluated only on the first 465 sequences due to time constraints.
    Results of the other trackers were taken from their papers and from~\cite{doersch2024bootstap} in case of \dataset{RoboTAP}.
    The \method{RoMa}-based correspondences chained by \method{MFTIQ} provide a very good position precision (\deltaavg) - best on \dataset{DAVIS}, second on \dataset{RoboTAP} and \dataset{Kinetics}.
    The occlusion accuracy (OA) is lower, also affecting the AJ score.
    The speed is compared with points-per-second (PPS). Timing computed on RTX\,A5000.
    Timing with $\dagger$ obtained from~\cite{doersch2023tapir} (TESLA\,V100), with $\ddagger$ from~\cite{li2024taptr} (TESLA\,A100), and with $*$ from \cite{cho2024flowtrack} (RTX\,3090) and recalculated to PPS. 
    }
  \label{tab:main-results}
\end{table*}

\paragraph{MFTIQ vs MFT Chain Selection}
We further evaluate the \method{MFTIQ} chain selection and how it compares to the original \method{MFT} on \dataset{TAP-Vid DAVIS}.
The~\cref{fig:mft-vs-mftiq-chaining} shows that the uncertainty score chaining of \method{MFT} leads to a significant preference of selecting short chains with big $\Delta$s.
In particular, the optical flow matching directly between the \template and the current frame ($\Delta = t - 1$) without chaining is selected with probability increasing with the current frame number.
However the probability of this selection being accurate decreases rapidly during the video.
On the other hand \method{MFTIQ} selects the short chains with big deltas conservatively, keeping the result accuracy high.
In other words, given the same optical flows, \method{MFTIQ} selects chains leading to better accuracy.

\subsection{Planar Object Tracking Dataset}
\label{subsec:plan-object-track}
In addition to the point-tracking benchmark, we have evaluated the proposed \method{MFTIQ} on the \dataset{POT-210}~\cite{liang2017planar} planar object tracking dataset.
The \dataset{POT-210} contains 210 videos capturing rigid flat objects.
The target is specified by coordinates of four control points forming a rectangle on the first frame of each video.
The tracker is to output the positions of these control points on each frame of the video.
There are 30 objects in total in \dataset{POT-210}, each captured in seven scenarios: \emph{motion blur}, \emph{occlusion}, \emph{out-of-view}, \emph{perspective distortion}, in-plane \emph{rotation}, \emph{scale change}, and \emph{unconstrained} combining all of the previous challenging factors.
From these only the partial occlusion factor is present in \dataset{TAP-Vid} point-tracking benchmark.

Since there is no occlusion ground-truth available on \dataset{POT-210}, we evaluate only the \deltaavg \dataset{TAP-Vid} metric.
Also we discard points outside the initialization region on the first frame as we only have ground-truth for the planar object.
We scale the output coordinates to $256 \times 256$ resolution as usual~\cite{doersch2022tap} and evaluate with the standard $1, 2, 4, 8, 16$ point error thresholds.
The results in \cref{tab:mftiq_pot_pointtracking} indicate overall good performance, with \method{RoMa}-based \method{MFTIQ} being particularly good on \emph{rotation} and \emph{scale change} scenarios compared to the plain \method{RoMa}.

\begin{table}
  \centering
  \setlength{\tabcolsep}{5pt}
  \begin{tabular}{lcc@{\hspace{1pt}}l}
    \toprule
                  & \method{RoMa}       & \multicolumn{2}{c}{MFTIQ}               \\
    challenge     & \deltaavg$\uparrow$ & \multicolumn{2}{c}{\deltaavg$\uparrow$} \\
    \midrule
    blur          & 88.4                & 86.9 & \resdeltan{-1.5}                 \\
    occlusion     & 99.2                & 96.9 & \resdeltan{-2.3}                 \\
    out-of-view   & 90.7                & 89.2 & \resdeltan{-1.5}                 \\
    perspective   & 96.8                & 94.8 & \resdeltan{-2.0}                 \\
    rotation      & 72.8                & 96.5 & \resdeltap{+23.7}                \\
    scale         & 92.2                & 98.0 & \resdeltap{+5.8}                 \\
    unconstrained & 93.5                & 93.3 & \resdeltan{-0.2}                 \\
    \midrule all  & 90.5                & 93.7 & \resdeltap{+3.2}                 \\
    \bottomrule
  \end{tabular}
  \caption{\method{MFTIQ RoMa} performance on POT-210 using a point-tracking metric, compared to plain \method{RoMa}.
  While the plain \method{RoMa} performs slightly better on some of the challenging scenarios, \method{MFTIQ} is significantly better on \emph{rotations} and \emph{scale change} due to the flow chaining, making it better on average -- \emph{all}.}
  \label{tab:mftiq_pot_pointtracking}
\end{table}

\noindent\textbf{MFTIQ Homography Tracking.} On top of the point-tracking evaluation, we also propose and evaluate a simple MFTIQ-based planar homography tracker.
We initialize MFTIQ on the initial frame and let it tracking all the initial frame pixels to get dense correspondences between the first and the current frame.
On each frame we mask out the background correspondences, \ie outside the initial rectangle on the first frame.
Finally we use the correspondences to robustly (with \method{RANSAC}~\cite{fischler1981random,barath2020magsacpp}) estimate a planar homography $\mat{H} \in \mathbb{R}^{3 \times 3}$ mapping from the initial to the current frame and transfer the control points from the initial frame into the current frame with $\mathbf{H}^{*}$ to get their current position.

This \method{MFTIQ RoMa} homography tracker out-performs the state-of-the-art on the \dataset{POT-210} benchmark as shown in~\cref{tab:MFTIQ_POT}.
The MFTIQ planar tracker performs particularly well on the \emph{blur} subset of POT-210, which contains many frames on which trackers fail due to big motion blur.
MFTIQ is able to recover from such failures by ``jumping'' over the problematic frames using the optical flows with bigger frame delta.
Note that the resulting planar tracker is only practical for real-time online application when the optical flows and IQ module features are precomputed.

\begin{table}
  \setlength{\tabcolsep}{5pt}
  \centering
  \begin{adjustbox}{max width=0.48\textwidth}
  \begin{tabular}{lrrrrrrr|r}\toprule
    method & BL & OCCL & OOV & PERS & ROT & SC & UNC & all\\
    \midrule
    \method{LISRD} \cite{pautrat2020online,liang2021planar} & 54.1 & 93.8 & 83.7 & 65.0 & 86.3 & 30.0 & 67.1 & 68.3\\
    \method{HDN} \cite{zhan2021homography} & 48.8 & 78.2 & 66.1 & 54.4 & 91.4 & 94.8 & 60.7 & 70.9\\
    \method{CGN} \cite{li2023centroid}& 41.6 & 88.1 & 82.8 & 76.5 & 96.1 & 90.3 & 72.4 & 78.5\\
    \method{WOFT} \cite{serych2023planar}& 60.4 & \frst{98.6} & 96.3 & 95.4 & 99.3 & 94.0 & 88.2 & 90.4\\
    \method{HVC-Net} \cite{zhang2022hvcnet}& 60.5 & \frst{98.6} & \frst{97.2} & 92.7 & 99.3 & 100.0 & \frst{90.1} & 91.4\\
    \textbf{\method{MFTIQ} (ours)} & \frst{72.0} & \frst{98.6} & 95.0 & \frst{96.6} & \frst{99.5} & \frst{100.0} & 89.1 & \frst{93.1}\\
    \bottomrule
  \end{tabular}
  \end{adjustbox}
  \caption{MFTIQ evaluation on planar tracking POT-210~\cite{liang2017planar} benchmark.
    Percentage of frames with alignment error under $5$px threshold evaluated on the improved ground-truth from~\cite{serych2023planar}.
  The RoMa-based MFTIQ followed by a RANSAC homography estimation on the resulting correspondences sets a new state-of-the-art performance.
    It achieves the most significant performance gain $+11.5\%$ on the \textit{BLur} sequences.
  }
  \label{tab:MFTIQ_POT}
\end{table}

\section{Conclusion}
In this work, we propose MFTIQ, a novel method for dense long-term tracking of points in video sequences.
By leveraging flow-chaining of the Multi-Flow Tracker (MFT) and enhancing it with our Independent Quality (IQ) module, MFTIQ significantly improves tracking accuracy and flexibility compared to existing methods.
Our approach effectively decouples the estimation of correspondence quality from the optical flow computations, enabling MFTIQ to handle complex occlusions and maintain accurate trajectories over extended period of time.

The ``plug-and-play'' nature of MFTIQ, which allows for seamless integration with any off-the-shelf optical flow method, further exemplifies its practical utility.
This flexibility enables users to choose the most suitable optical flow method based on their specific performance or computational efficiency needs without the requirement for additional fine-tuning or architectural adjustments.

Our experimental results demonstrate that MFTIQ not only matches but often surpasses current state-of-the-art point-tracking methods in terms of accuracy and speed.
Particularly, it shows significant improvements over the baseline MFT method.
%, especially in challenging scenarios involving rapid movements and prolonged occlusions.
The capability of MFTIQ to operate efficiently with various optical flow methods underscores its robustness and adaptability.
Looking forward, we anticipate that future advancements in optical flow technology will further enhance the performance of MFTIQ\@.
The architecture's compatibility with evolving flow estimation techniques promises continual improvements in tracking precision and computational efficiency.
We publish\footnote{\url{https://github.com/serycjon/MFTIQ}} the MFTIQ code and models.

{\bf Acknowledgements.}
This work was supported by Toyota Motor Europe,
and by the Grant Agency of the Czech Technical University in Prague, grant No.SGS23/173/OHK3/3T/13.

%%%%%%%%% REFERENCES
{\small
\bibliographystyle{wacvTemplate/ieee_fullname}
\bibliography{sjo-bib,jabref}
}

\clearpage
\appendix
\newpage
{\vskip .375in}
\begin{center}
  % smaller title font only for rebuttal
  {\Large \textbf{Supplementary Materials}}\\
  \vspace*{8pt}
\end{center}

% https://tex.stackexchange.com/a/3529
\begin{table*}[b]
  \centering
  \begin{tabular}{l|rrrr|rrrr}
    \toprule
    & \multicolumn{2}{c}{FPS $\uparrow$} & \multicolumn{2}{c}{PPS $\uparrow$} & \multicolumn{2}{|c}{FPS pre-computed $\uparrow$} & \multicolumn{2}{c}{PPS pre-computed $\uparrow$} \\
    \method{MFTIQ} with                           & 
                                                    \multicolumn{1}{l}{$512 \stimes 512$}                 & 
                                                                                                            \multicolumn{1}{l}{$720 \stimes 1080$}                & 
                                                                                                                                                                    \multicolumn{1}{l}{$512 \stimes 512$}                 & 
                                                                                                                                                                                                                            \multicolumn{1}{l}{$720 \stimes 1080$}                & 
                                                                                                                                                                                                                                                                                    \multicolumn{1}{|l}{$512 \stimes 512$}                & 
                                                                                                                                                                                                                                                                                                                                            \multicolumn{1}{l}{$720 \stimes 1080$}                & 
                                                                                                                                                                                                                                                                                                                                                                                                    \multicolumn{1}{l}{$512 \stimes 512$}                 & 
                                                                                                                                                                                                                                                                                                                                                                                                                                                            \multicolumn{1}{l}{$720 \stimes 1080$} \\ \midrule
    \method{RAFT}~\cite{teed2020raft}             & 2.66 & 0.90                        & 8234  & 8897                       & 10.95 & 3.76                              & 26944 & 33921 \\
    % \method{NeuFlow}~\cite{zhang2024neuflow}      & 6.01 & 2.08                        & 17621 & 18546                      & 10.96 & 3.68                              & 27438 & 32396 \\
    \method{NeuFlowV2}~\cite{Zhang2024a}      & 5.67         & 2.03               & 16446         & 19348      & 10.56   & 3.59         & 27589               & 32175                                \\
    \method{RAPIDFlow}~\cite{Morimitsu2024}      & 3.06         & 1.35               & 9603         & 13058      & 10.65    & 3.49         & 28396               & 31960                                  \\
    \method{GMFlow}~\cite{Xu2022b}      & 3.63         & 0.76               & 11365         & 7638      & 10.31    & 3.47         & 27304               & 32075                                  \\
    \method{SEA-RAFT}~\cite{Wang2024b}      & 2.93         & 0.93               & 9285         & 9195      & 10.24   & 3.40         & 27296               & 31591                                   \\
    % \hspace{0.5em}\method{RPKNet}~\cite{Morimitsu2024a}      & 62.78         & 76.61               & 86.39         & XX      & XX                                     \\
    % \hspace{0.5em}\method{LLA-Flow}~\cite{Xu2023}      & 61.78         & 75.18               & 85.44         & XX      & XX  
    \method{MemFlow}~\cite{dong2024memflow}       & 1.16 & 0.29                        & 3836  & 2985                       & 10.95 & 3.71                              & 27412 & 32907 \\
    \method{FFormer++}~\cite{shi2023flowformerpp} & 1.04 & 0.24                        & 3457  & 2437                       & 10.47 & 3.76                              & 27183 & 33303 \\
    \method{RoMa}~\cite{edstedt2023roma}          & 0.21 & 0.19                        & 709   & 1948                       & 10.10 & 3.67                              & 24986 & 32703 \\
    \bottomrule
  \end{tabular}
  \caption{\textbf{Runtime evaluation} of the whole \method{MFTIQ} tracker with various OF methods with \textit{(right)} and without \textit{(left)} OF and features pre-computed.
    All results shows processing speed in frames-per-second (FPS) and points-per-second (PPS) for two different resolutions of images.
    PPS were evaluated for a sequence of 80 images.
    In the case of pre-computed optical flow and image feature cache, speed is the same regardless of the OF method used up to a measurement noise.
  }
  \label{tab:MFTIQ_OF_TIMING}
\end{table*}

\section{Image Feature Extraction}
\label{sec:image-feat-extraction}
For the \method{DinoV2} features we use the author-provided \texttt{ViT-S/14-reg} network checkpoint.
The ResNet50~\cite{he15resnet} network, pre-trained on the ImageNet1K~\cite{deng09imagenet} dataset, is used to extract features from its first three blocks: the input block, residual block 1, and residual block 2.
Each output feature is up-sampled to $\frac{H}{4} \times \frac{W}{4}$ and compressed to 32 channels using a convolutional layer.

The custom image features CNN is trained from scratch, and it is inspired by \method{NeuFlow}'s feature CNN~\cite{zhang2024neuflow}.
Initially, an image pyramid is created by subsampling the input image at different scales (1/1, 1/2, 1/4).
For each level of the image pyramid, a convolutional layer is applied with specific \underline{k}ernel sizes, \underline{s}trides, and \underline{p}adding to ensure the output resolution is $\frac{H}{4} \times \frac{W}{4}$ (k4:s4:p0\,$\vert$\,k8:s2:p3\,$\vert$\,k7:s1:p3).
The outputs from each pyramid level are concatenated and compressed to 32 channels using an additional convolutional layer.

The features from all the feature providers (\method{DinoV2}, \method{ResNet}, custom CNN) are aggregated and compressed through a convolutional operation (from $5\times32$ channels down to 32 channels) to produce an additional \emph{fused feature} for the cost-volume.

The impact of feature extractors on performance is demonstrated in~\cref{tab:MFTIQ_ABLATION_UOM}.
Excluding \method{DinoV2} features causes a decrease in AJ from $65.7$ to $64.6$. Further removing both \method{DinoV2} and \method{ResNet} features, leaving only the custom shallow CNN features, results in a more pronounced drop to AJ $61.5$.
Since the overall runtime is dominated by optical flow computation, it remains nearly unchanged ($\approx -0.01$~FPS) without the \method{Dino} and the \method{ResNet} backbones. Thus, we keep all three feature extractors.

\begin{figure}
  \centering
  \includegraphics[width=0.8\linewidth,trim={0 0.4cm 0 0.25cm},clip]{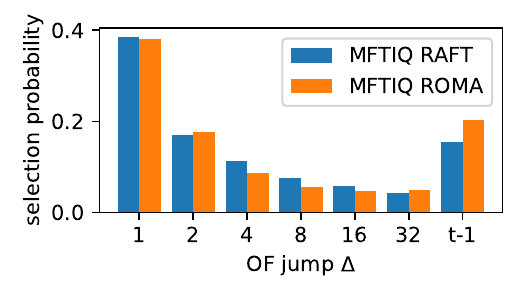}
  \vspace{-1ex}
  \caption{Probability of selecting OF with a given $\Delta$ on \method{TAP-Vid DAVIS}~\cite{doersch2022tap}, evaluated on frames more than 32 frames distant from template.
    Statistics are similar for \method{RAFT} and \method{RoMa}, but long jumps $\Delta = t-1$ are selected more often with \method{RoMa}.
    }
  \label{fig:delta-statistics}
\end{figure}

\section{Feature Concatenation and Flow Features}
\label{sec:feat-aggr-after}
The final featuremap contains 6 (\method{Dino}, $3\times$ \method{ResNet}, custom CNN, \emph{fused}) cost-volumes, each flattened to 49 channels from the $\pm 3$ range $7\times 7$ cost-volume response maps, resulting in a total of 294 channels.
In addition to that it contains $2\times 32$ channels of the \emph{fused features} from the \template and the current frame (warped by the flow).
Finally it has 64 channels of flow features derived from the input $\Flow_{\ft{1}{t}}$ flow chain by a small CNN, for a grand total of 422 channels.

\section{Timing}
\label{sec:timing}
As mentioned in Section~\ref{subsec:implementation-details}, we implement caching for optical flow estimates and image features to improve efficiency.
Table~\ref{tab:MFTIQ_OF_TIMING} reports the overall tracking timing for results shown in \cref{tab:MFTIQ_OF} and \cref{tab:main-results} in the paper both with standard caching during computation and with the caches pre-computed offline.
With optical flow and image features computed in advance, \method{MFTIQ} runs at $3.7$ FPS on $720\times 1080$ and at over $10$ FPS on $512\stimes 512$ video resolution.

\begin{table*}
  \setlength{\tabcolsep}{5pt}
  \centering
  \begin{tabular}{lccc}
  \toprule
                                                                
      method                                             & AJ $\uparrow$ & \deltaavg$\uparrow$ & OA $\uparrow$ \\
    \midrule
    (1)\hspace{0.5em}Full \method{MFTIQ} (\method{RoMa}) & 65.67         & 79.82               & 87.75         \\
    (2)\hspace{0.5em} -\method{Dino}                     & 64.61         & 79.59               & 87.80         \\
    (3)\hspace{0.5em} -\method{Dino} -\method{ResNet}    & 61.54         & 78.58               & 85.02         \\
  \bottomrule
  \end{tabular}
  \caption{Influence of IQ feature extractors in the \method{MFTIQ} model. The table shows the performance variations when different backbones are omitted, with the remainder of the network held constant. All models followed identical training and evaluation protocols.
  The evaluation was conducted using the \dataset{TAP-Vid DAVIS}~\cite{doersch2022tap} (strided) dataset.
  }
  \label{tab:MFTIQ_ABLATION_UOM}
\end{table*}

\begin{table*}
  \setlength{\tabcolsep}{5pt}
  \centering
%   \begin{adjustbox}{max width=0.48\textwidth}
  \begin{tabular}{lcccrr}
  \toprule
                                                                &               &                     &               & \multicolumn{2}{c}{runtime [FPS] $\downarrow$} \\
      $\Delta$-set hyper-parameter                                                    & AJ $\uparrow$ & \deltaavg$\uparrow$ & OA $\uparrow$ & 512x512 & 720x1080                               \\
    \midrule
   % \midrule
    $\Delta \in \{1,2,4,8,16,32,t-1\}$                          & 65.67         & 79.82               & 87.75         & 0.21      & 0.19                                    \\
    $\Delta \in \{1,4,16,t-1\}$      & 65.50	& 79.57 &	87.42         & 0.35      & 0.32                                     \\
    $\Delta \in \{1,8,32,t-1\}$      & 59.03	& 72.79 &	82.34         & 0.35      & 0.32                                     \\
    $\Delta \in \{t-1\}$      & 57.46	& 70.08 &	78.73         & 1.31      & 1.14                                     \\
    $\Delta \in \{1\}$      & 54.67	& 70.99 &	73.35         & 1.31      & 1.14                                    \\
    % 
    % \rowcolor{blue!10}%
  \bottomrule
  \end{tabular}
%   \end{adjustbox}
  \caption{Ablation of different sets of $\Delta$ used for optical flow chaining.
    The default set of $\Delta$s \textit{(first row)} (same as in \method{MFT}) performs the best.
    The base-4 \textit{(second row)} set achieves a better speed / performance trade-off.
  \method{MFTIQ RoMa} evaluated on \dataset{TAP-Vid DAVIS}~\cite{doersch2022tap} (strided).
  Performance measured by average Jaccard (AJ), position accuracy (\deltaavg), and occlusion accuracy (OA). Speed of tracking densely measured by average frames per second (FPS).}
  \label{tab:MFTIQ_ABLATION_STEPS}
\end{table*}

\section{Delta Set Ablation}
\label{sec:delta-value-ablation}
\cref{tab:MFTIQ_ABLATION_STEPS} shows the effect of using different sets of $\Delta$s. Our default base-2 configuration, $\Delta \in \left\{1, 2, 4, 8, 16, 32, t-1\right\}$, follows the MFT setup~\cite{neoral2024mft}. However, we found that using a base-4 set, $\Delta \in \left\{ 1, 4, 16, t-1 \right\}$, achieves a $1.6\times$ speedup with only a minimal performance decrease on the \method{TAP-Vid DAVIS} dataset~\cite{doersch2022tap}.
Both direct matching between the template and the current frame ($\Delta \in \left\{ t - 1 \right\}$) and consecutive frame chaining ($\Delta \in \left\{ 1 \right\}$) result in a significant performance decrease across all evaluated metrics.

We have also evaluated (\cref{fig:delta-statistics}) the frequency of selection for each $\Delta$ in \method{MFTIQ RAFT} and \method{MFTIQ RoMa} in the default $\Delta$-set.
The results show similar statistics between the two OFs, though the direct jump ($\Delta = t - 1$) is selected more frequently in \method{RoMa}.
This is expected since the \method{RoMa} was trained on wide-baseline matching data, making it more reliable with more distant pairs of frames.
Only frames beyond timestep 32 are evaluated to avoid biasing the results with smaller $\Delta$s at the beginning of the sequence, where longer $\Delta$s are not yet available for matching.

\end{document}